\documentclass[twocolumn, 12pt]{article}
\usepackage[english]{babel}
\usepackage{booktabs}
\usepackage{color, colortbl}
\usepackage[labelfont=bf]{caption}
\usepackage[cm]{fullpage}
\usepackage{graphicx}
\usepackage[utf8]{inputenc}
\usepackage[numbers]{natbib}
\usepackage{parskip}
\usepackage{subcaption}

\PassOptionsToPackage{hyphens}{url}
\usepackage[
  colorlinks=true,
  linkcolor=black,
  anchorcolor=black,
  citecolor=black,
  filecolor=black,
  menucolor=black,
  runcolor=black,
  urlcolor=black]{hyperref}

\definecolor{Gray}{gray}{0.9}
\title{The History of Speech Recognition to the Year 2030}
\author{Awni Hannun\footnote{
  Send correspondence to
  \href{mailto:awni.hannun@gmail.com}{awni.hannun@gmail.com}}}

\date{\today}

\begin{document}
\maketitle

\begin{abstract}
    The decade from 2010 to 2020 saw remarkable improvements in automatic
    speech recognition. Many people now use speech recognition on a daily
    basis, for example to perform voice search queries, send text messages, and
    interact with voice assistants like Amazon Alexa and Siri by Apple. Before
    2010 most people rarely used speech recognition. Given the remarkable
    changes in the state of speech recognition over the previous decade, what
    can we expect over the coming decade? I attempt to forecast the state of
    speech recognition research and applications by the year 2030. While the
    changes to general speech recognition accuracy will not be as dramatic as
    in the previous decade, I suggest we have an exciting decade of progress in
    speech technology ahead of us.
\end{abstract}

\section{Recap}
\label{sec:recap}

\begin{figure*}
\centering
\includegraphics[width=\linewidth]{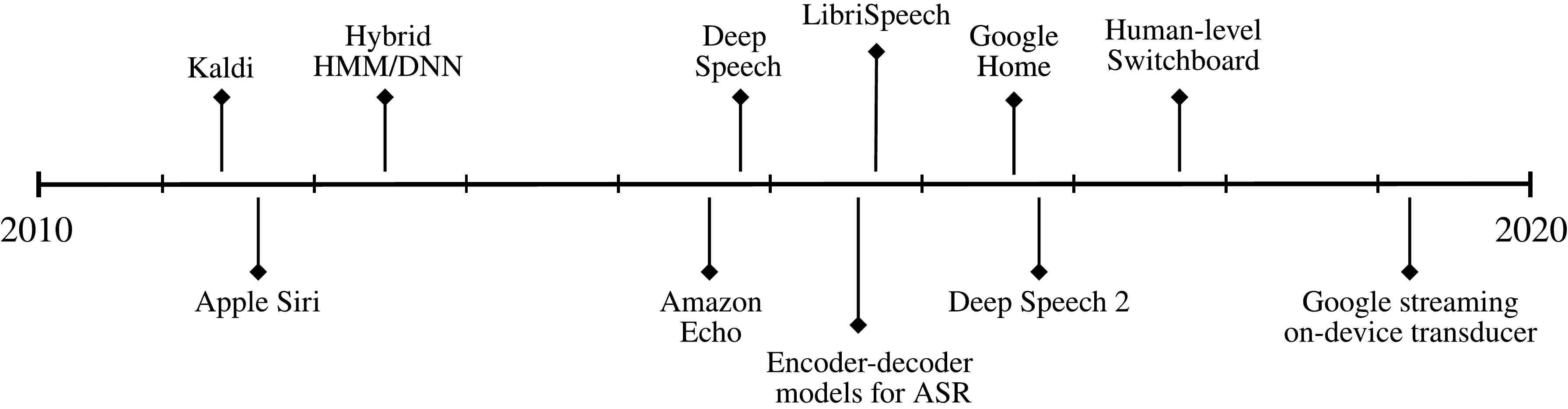}
\caption{A timeline of some of the major developments in speech recognition
    from the years 2010 to 2020. The decade saw the launch of
    voice-based devices and voice assistants, open-source and widely used
    speech recognition software like Kaldi~\citep{povey2011kaldi}, and larger
    benchmarks like LibriSpeech~\citep{panayotov2015librispeech}. We also saw
    speech recognition models improve starting from hybrid neural network
    architectures~\citep{hinton2012deep} to more end-to-end models including
    Deep Speech~\citep{hannun2014deep}, Deep Speech 2~\citep{amodei2016deep},
    encoder-decoder models with attention~\citep{chorowski2015attention}, and
    transducer-based speech recognition~\citep{he2019streaming}.}
\label{fig:asr_timeline}
\end{figure*}

The decade from 2010 to 2020 saw remarkable progress in speech recognition and
related technology. Figure~\ref{fig:asr_timeline} is a timeline of some of the
major developments in the research, software, and application of speech
recognition over the previous decade. The decade saw the launch and spread of
phone-based voice assistants like Apple Siri. Far-field devices like Amazon
Alexa and Google Home were also released and proliferated.

These technologies were enabled in-part by the remarkable improvement in the
word error rates of automatic speech recognition as a result of the rise of
deep learning. The key drivers of the success of deep learning in speech
recognition have been 1) the curation of massive transcribed data sets, 2) the
rapid rate of progress in graphics processing units, and 3) the improvement in
the learning algorithms and model architectures.

Thanks to these ingredients, the word error rate of speech recognizers improved
consistently and substantially throughout the decade. On two of the most
commonly studied benchmarks, automatic speech recognition word error rates have
surpassed those of professional transcribers (see figure~\ref{fig:wers}).

This remarkable progress invites the question: what is left for the coming
decade to the year 2030? In the following, I attempt to answer this question.
But, before I begin, I'd first like to share some observations on the general
problem of predicting the future. These findings are inspired by the
mathematician (as well as computer scientist and electrical engineer) Richard
Hamming, who also happened to be particularly adept at forecasting the future
of computing.

\section{On Predicting the Future}
\label{sec:predicting_future}

Richard Hamming in \emph{The Art of
Doing Science and Engineering}~\citep{hamming1997art} makes many predictions,
many of which have come to pass. Here are a few examples\footnote{Quotes and
predictions are from chapters 2, 4, and 21 of \citet{hamming1997art}.}
\begin{itemize}
    \item He stated that by ``the year 2020 it would be fairly universal
        practice for the expert in the field of application to do the actual
        program preparation rather than have experts in computers (and ignorant
        of the field of application) do the program preparation.''
    \item He predicted that neural networks ``represent a solution to the
        programming problem,'' and that ``they will probably play a large part
        in the future of computers.''
    \item He predicted the prevalence of general-purpose rather than
        special-purpose hardware, digital over analog, and high-level
        programming languages all long before the field had decided one way or
        another.
    \item He anticipated the use of fiber-optic cables in place of copper wire
        for communication well before the switch actually took place.
\end{itemize}

These are just a few examples of Hamming's extraordinary prescience. Why was he
so good at predicting the future? Here are a few observations which I think
were key to his ability:

{\bf Practice:} One doesn't get good at predicting the future without actually
practicing at it. Hamming practiced. He reserved every Friday afternoon ``great
thoughts'' in which he mused on the future. But he didn't just predict in
isolation. He made his predictions public, which forced him to put them in a
communicable form and held him accountable. For example, in 1960 Hamming gave a
talk titled ``The History of Computing to the Year 2000'' (you may recognize
the title) in which he predicted how computing would evolve over the next
several decades.

{\bf Focus on fundamentals:} In some ways, forecasting the future development
of technology is just about understanding the present state of technology more
than those around you. This requires both depth in one field as well as
non-trivial breadth. This also requires the ability to rapidly assimilate new
knowledge. Mastering the fundamentals builds a strong foundation for both.

{\bf Open mind:} Probably the most important trait Hamming exhibited, and in my
opinion the most difficult to learn, was his ability to keep an open mind.
Keeping an open mind requires constant self-vigilance. Having an open mind one
day does not guarantee having it the next. Having an open mind with respect to
one scientific field does not guarantee having it with respect to another.
Hamming recognized for example that one may be more productive in an office
with the door closed, but he kept his office door open as he believed an ``open
mind leads to the open door, and the open door tends to lead to the open
mind''~\citep[chp. 30]{hamming1997art}.

I'll add to these a few more thoughts. First, the rate of change of progress in
computing and machine learning is increasing. This makes it harder to predict
the distant future today than it was 50 or 100 years ago. These days predicting
the evolution of speech recognition even ten years out strikes me as a
challenge. Hence my choosing to work with that time frame.

\begin{figure}
    \centering
    \includegraphics[width=\linewidth]{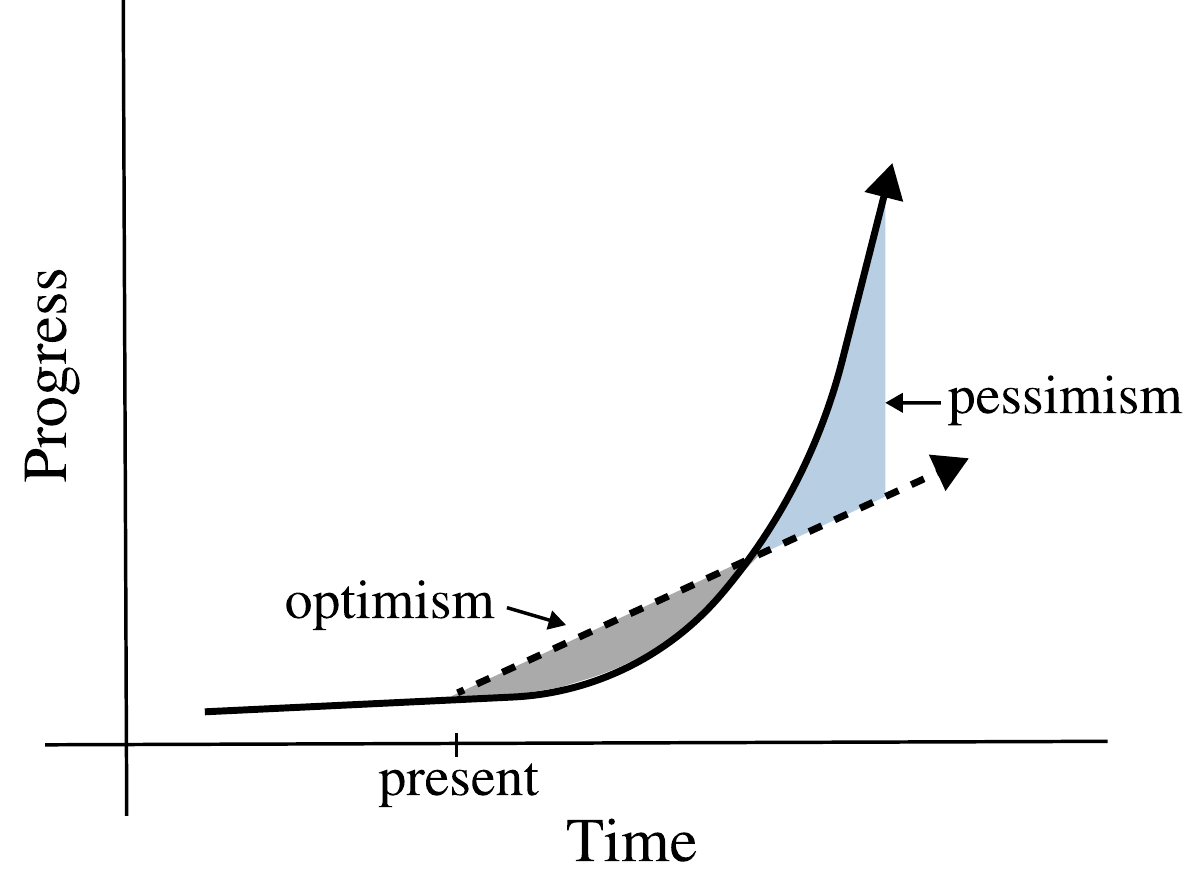}
    \caption{The graph depicts progress as a function of time. The linear
    extrapolation from the present (dashed line) initially results in overly
    optimistic predictions.  However, eventually the predictions become
    pessimistic as they are outstripped by the exponential growth (solid
    line).}
    \label{fig:exponential_growth}
\end{figure}

A common saying about technology forecasting is that short-term predictions
tend to be overly optimistic and long-term predictions tend to be overly
pessimistic. This is often attributed to the fact that progress in technology
has been exponential. Figure~\ref{fig:exponential_growth} shows how this
happens if we optimistically extrapolate from the present assuming progress is
linear with time. Progress in speech recognition over the previous decade
(2010-2020) was driven by exponential growth along two key axes. These
were compute (e.g. floating-point operations per second) and data set
sizes. Whether or not figure~\ref{fig:exponential_growth} applies to speech
recognition for the coming decade remains to be seen.

I'm sure a lot of the following predictions will prove wrong. In some ways,
particularly when it comes to the more controversial predictions, these are
really more of an optimistic wishlist for the future. On that note, let me
close this section with the famous words of the computer scientist Alan
Kay\footnote{Alan Kay is best known for developing the modern graphical user
interface and also object-oriented programming in the Smalltalk programming
langauge.}:
\begin{quote}
    \emph{The best way to predict the future is to invent it.}
\end{quote}

\section{Research Predictions}
\label{sec:research_predictions}

\subsection{Semi-supervised Learning}

{\bf Prediction:} Semi-supervised learning is here to stay. In particular,
self-supervised pretrained models will be a part of many machine-learning
applications, including speech recognition.

Part of my job as a research scientist is hiring, which means a lot of
interviews. I've interviewed more than a hundred candidates working on a
diverse array of machine-learning applications. Some large fraction,
particularly of the natural language applications, rely on a pretrained model
as the basis for their machine-learning enabled product or feature.
Self-supervised pretraining is already pervasive in language applications in
industry. I predict that by 2030 self-supervised pretraining will be just as
pervasive in speech recognition.

The past three years of deep learning have been the years of semi and
self-supervision. The field has undoubtedly learned how to improve
machin-learning models using unannotated data. Self-supervised
learning~\cite{lecun2021self} has benefited many of the most challenging
machine learning benchmarks. In language tasks, state-of-the-art records have
been repeatedly set and surpassed by self-supervised
models~\citep{devlin2019bert, radford2019language, yang2019xlnet}. Self and
semi-supervision are now commonplace and setting records in computer
vision~\citep{he2020momentum, chen2020simple, grill2020bootstrap}, abstractive
summarization~\citep{zhang2020pegasus} and machine
translation~\citep{sennrich2016improving}.

Speech recognition has also benefited from semi-supervised learning. Two
approaches are commonly used, both of which work well. The first approach is
self-supervised pretraining~\citep{schneider2019wav2vec, zhang2020pushing} with
a loss function based on contrastive predictive
coding~\citep{oord2018representation}. The idea is simple: train the model to
predict the future frame(s) of audio given the past. Of course, the devil is in
the details and the scale. The second approach is
pseudo-labeling~\citep{lee2013pseudo, kahn2020self, xu2020iterative}. Again the
idea is simple: use the trained model to predict the label on unlabeled data,
then train a new model on the predicted labels as if they were the ground
truth. And again the devil is in the details and the scale. The fact
that pseudo-labeling leads to better models is remarkable. It feels as if we
are getting something for nothing, a free lunch. The reason and the regime in
which pseudo-labeling works are interesting research questions.

The main challenges with self-supervision are those of scale, and hence
accessibility. Right now only the most highly endowed industry research labs
(\emph{e.g.} Google Brain, Google DeepMind, Facebook AI Research, OpenAI,
\emph{etc.}) have the funds to burn on the compute required to research
self-supervision at scale. As a research direction, self-supervision is only
becoming less accessible to academia and smaller industry labs.

{\bf Research implications:} Self-supervised learning would be more accessible
given lighter-weight models which could be trained efficiently on less data.
Research directions which could lead to this include sparsity for
lighter-weight models, optimization for faster training, and effective ways of
incorporating prior knowledge for sample efficiency.

\begin{figure*}[ht!]
    \centering
    \begin{subfigure}[b]{0.48\textwidth}
    \centering
    \includegraphics[width=\linewidth]{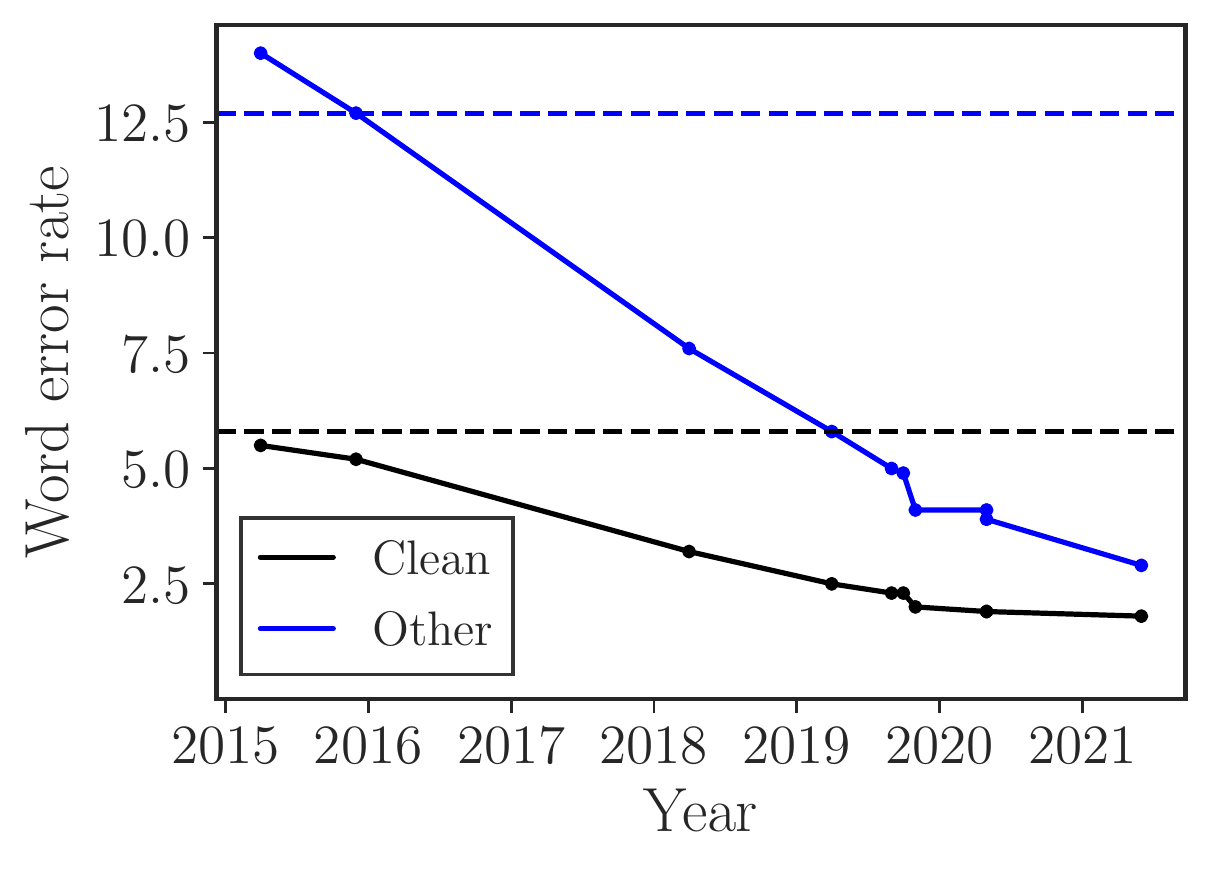}
    \caption{LibriSpeech}
    \end{subfigure}
    \hfill
    \begin{subfigure}[b]{0.48\textwidth}
    \centering
    \includegraphics[width=\linewidth]{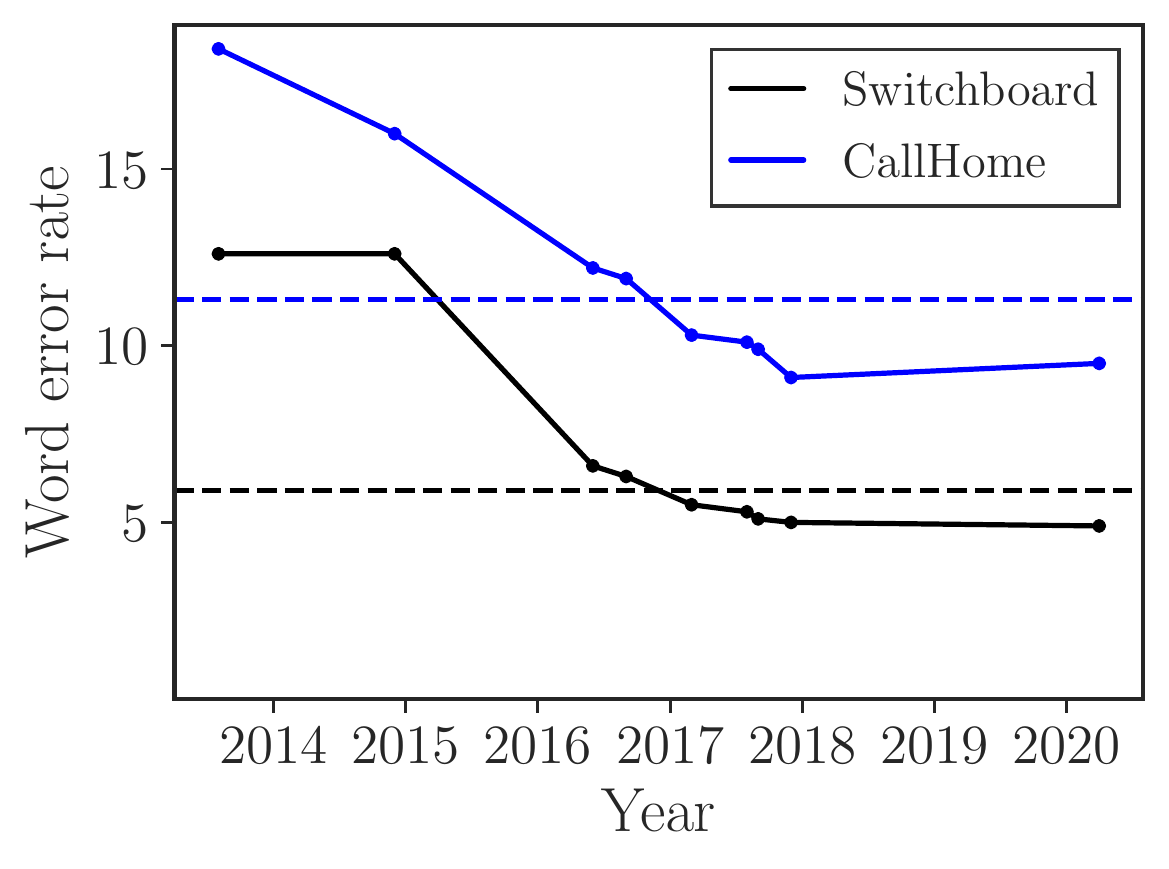}
    \caption{Switchboard Hub5'00}
    \end{subfigure}
    \caption{The improvement in word error rate over time on the
    LibriSpeech~\citep{panayotov2015librispeech} and Switchboard Hub5'00
    benchmarks. The data for these figures is from
    \url{https://github.com/syhw/wer_are_we}. The dashed lines indicate
    human-level performance. The human-level results on LibriSpeech are
    reported in \citet{amodei2016deep}, and those on Switchboard are reported in
    \citet{xiong2016achieving}.}
    \label{fig:wers}
\end{figure*}

\subsection{On Device}
\label{sec:on_device}

{\bf Prediction:} Most speech recognition will happen on the device or at the
edge.

There are a few reasons I predict this will happen. First, keeping your data on
your device rather than sending it to a central server is more private. The
trend towards data privacy will encourage on-device inference whenever
possible. If the model needs to learn from a user's data, then the training
should happen on the device.

The second reason to prefer on-device inference is latency. In absolute terms,
the difference between 10 milliseconds and 100 milliseconds is not much.  But
the former is well below the perceptual latency of a human, and the latter well
above~\citep{lago2004quest, levitin2000perception}.  Google has already
demonstrated an on-device speech recognition system with accuracy nearly as
good as a server-side system~\citep{he2019streaming}. The latency differences
are easily noticeable.\footnote{For an example of the perceptual difference in
latencies see the blog post on Google's on-device speech recognizer:
\url{https://ai.googleblog.com/2019/03/an-all-neural-on-device-speech.html}}
From a pragmatic standpoint, the latency of the server-side recognizer is
probably sufficient. However, the imperceptible latency of the on-device system
makes the interaction with the device feel much more responsive and hence more
engaging.

A final reason to prefer on-device inference is 100\% availability. Having the
recognizer work even without an internet connection or in spotty service means
it will work all the time. From a user interaction standpoint there is a big
difference between a product which works most of the time and a product which
works every time.

{\bf Research implications:} On-device recognition requires models with smaller
compute and memory requirements and which use less energy in order to preserve
battery life. Model quantization and knowledge distillation (training a smaller
model to mimic the predictions of a more accurate larger model) are two
commonly used techniques. Sparsity, which is less commonly used, is another
approach to generate lighter-weight models. In sparse models, most of the
parameters (\emph{i.e.} connections between hidden states) are zero and can be
effectively ignored. Of these three techniques, I think sparsity is the
most promising research direction.

I believe we have extracted most of the value that quantization has to offer.
Even in the unlikely best possible scenario of further reducing quantization
from 8-bit to 1-bit, we only get a factor-of-eight gain. With distillation, we
still have a lot to learn. However, I believe uncovering the mechanism through
which distillation works will subsequently enable us to train small models
directly rather than taking the circuitous path of training a large model and
then a second small model to mimic the large model.

This leaves sparsity as the most promising research direction for
lighter-weight models. As findings like the ``lottery ticket hypothesis''
demonstrate~\citep{frankle2018lottery}, we have a lot to learn about the role
of sparsity in deep learning. In theory, the computational gains from sparsity
could be substantial. Realizing these gains will require developments in the
software, and possibly hardware, used to evaluate sparse models.

Weak supervision will be an important research direction for on-device training
for applications which typically require labeled data. For example, a users
interaction with the output of a speech recognizer or the actions they take
immediately afterward could be useful signal from which the model can learn in
a weakly-supervised manner.

\subsection{Word Error Rate}
\label{sec:wer}

{\bf Prediction:} By the end of the decade, possibly much sooner, researchers
will no longer be publishing papers which amount to ``improved word error rate
on benchmark X with model architecture Y.''

As you can see in figure~\ref{fig:wers}, word error rates on the two most
commonly studied speech recognition benchmarks have saturated. Part of the
problem is that we need harder benchmarks for researchers to study.  Two
recently released benchmarks may spur further research in speech
recognition~\citep{chen2021gigaspeech, galvez2021people}. However, I predict
that these benchmarks will quickly saturate by scaling up models and
computation.

Another part of the problem is that we have reached a regime where word error
rate improvements on academic benchmarks no longer correlate with practical
value. Speech recognition word error rates on both benchmarks in
figure~\ref{fig:wers} surpassed human word error rates several years
ago.\footnote{Estimates of human-level word error rates on the CallHome portion
of Hub5'00 vary considerably. For example \citet{saon2017english} report a best
result 6.8 out of three transcribers whose results vary by nearly 2.0 absolute
word error rate.} However, in most settings humans understand speech better
than machines do. This implies that word error rate as a measure of the quality
our speech recognition systems does not correlate well with an ability to
understand human speech.

A final issue is research in state-of-the-art speech recognition is becoming
less accessible as models and data sets are getting larger, and as computing
costs are increasing. A few well-funded industry labs are rapidly becoming the
only places that can afford this type of research. As the advances become more
incremental and further from academia, this part of the field will continue to
shift from research labs to engineering organizations.

\subsection{Richer Representations}

\begin{figure}
\centering
\includegraphics[width=\linewidth]{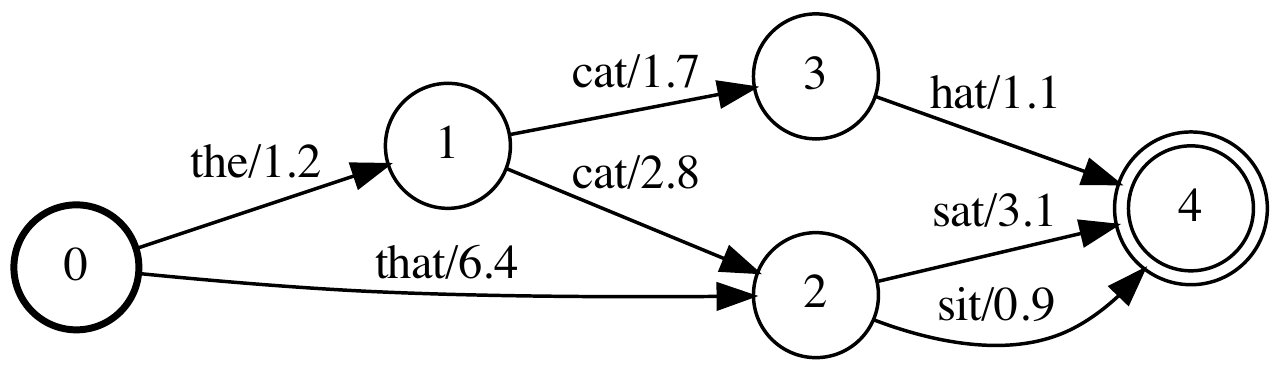}
\caption{An example lattice used to encode mutliple hypotheses output from a
    speech recognizer with differing weights.}
\label{fig:lattice}
\end{figure}

{\bf Prediction:} Transcriptions will be replaced by richer representations for
downstream tasks which rely on the output of a speech recognizer. Examples of
such downstream applications include conversational agents, voice-based search
queries, and digital assistants.

Downstream applications often don't care about a verbatim transcription; they
care about semantic correctness. Hence, improving the word error rate of a
speech recognizer often does not improve the objective of the downstream task.
One possibility is to develop a \emph{semantic error rate} and use
it to measure the quality of the speech recognizer. This is a challenging
albeit interesting research problem.

I think a more likely outcome is to give downstream applications richer
representations from the speech recognizer. For example, instead of passing a
single transcription, passing a lattice of possibilities (as in
figure~\ref{fig:lattice}) which captures the uncertainty for each could be much
more useful.

{\bf Research implications:} The exact structure used to encode the
representation is an open question. One possibility could be some sort of
weighted transducer which if differentiable could allow for fine-tuning the
recognizer to specific applications~\cite{k2, hannun2020differentiable}. This
type of representation also requires models which are able to ingest
variable-sized graphs as input.

\subsection{Personalization}

{\bf Prediction:} By the end of the decade, speech recognition models will be
deeply personalized to individual users.

One of the main distinctions between the automatic recognition of speech and
the human interpretation of speech is in the use of context. Humans rely on a
lot of context when speaking to one another. This context includes the topic
of conversation, what was said in the past, the noise background, and visual
cues like lip movement and facial expressions. We have, or will soon
reach, the Bayes error rate for speech recognition on short (\emph{i.e.} less
than ten second long) utterances taken out of context. Our models are using the
signal available in the data to the best of their ability. To continue to
improve the machine understanding of human speech will require leveraging
context as a deeper part of the recognition process.

One way to do this is with personalization. Personalization is already used to
improve the recognition of utterances of the form ``call \texttt{<NAME>}''.
\citet{sim2019personalization} found personalizing a model with a user's
contact list improves named entity recall from 2.4\% to 73.5\% -- a massive
improvement. Personalizing models to individual users with speech disorders
improves word error rates by 64\% relative~\citep{sim2019investigation}.
Personalization can make a huge difference in the quality of the
recognition, particularly for groups or domains that are underrepresented in
the training data. I predict we will see much more pervasive personalization by
the end of the decade.

{\bf Research implications:} On-device personalization requires on-device
training which in itself requires lighter-weight models and some form of weak
supervision (see section~\ref{sec:on_device}).  Personalization requires models
which can be easily customized to a given user or context. The best way to
incorporate such context into a model is still a research question.

\begin{table*}[ht!]
    \caption{Predictions for the progress in speech recognition research and applications
    by the year 2030.}
    \centering
    \begin{tabular}{l}
    \toprule
    Prediction \\
    \midrule
    Self-supervised learning and pretrained models are here to stay. \\
    \rowcolor{Gray} Most speech recognition (inference) will happen at the edge. \\
    On-device model training will be much more common. \\
    \rowcolor{Gray} Sparsity will be a key research direction to enable on-device inference and training. \\
    Improving word error rate on common benchmarks will fizzle out as a research goal. \\
    \rowcolor{Gray} Speech recognizers will output richer representations (graphs) for use by downstream tasks. \\
    Personalized models will be commonplace. \\
    \rowcolor{Gray} Most transcription services will be automated. \\
    Voice assistants will continue to improve, but incrementally. \\
    \bottomrule
    \end{tabular}
    \label{tab:predictions}
\end{table*}

\section{Application Predictions}
\label{sec:application_predictions}

\subsection{Transcription Services}

{\bf Prediction:} By the end of the decade, 99\% of transcribed speech services
will be done by automatic speech recognition. Human transcribers will perform
quality control and correct or transcribe the more difficult utterances.
Transcription services include, for example, captioning video, transcribing
interviews, and transcribing lectures or speeches.

\subsection{Voice Assistants}

{\bf Prediction:} Voice assistants will get better, but incrementally, not
fundamentally. Speech recognition is no longer the bottleneck to better voice
assistants. The bottlenecks are now fully in the language understanding domain
including the ability to maintain conversations, multi-ply contextual
responses, and much wider domain question and answering. We will continue to
make incremental progress on these so-called AI-complete
problems,\footnote{\citet[sec. 4]{shapiro1992encyclopedia} defines an AI task as
AI-complete if solving it is equivalent to ``solving the general AI problem'',
which he defines as ``producing a generally intelligent computer program''.}
but I don't expect them to be solved by 2030.

Will we live in smart homes that are always listening and can respond to our
every vocal beck and call? No. Will we wear augmented reality glasses on our
faces and control them with our voice? Not by 2030.

\section{Conclusion}
\label{sec:conclusion}

Table~\ref{tab:predictions} summarizes my predictions for the progress in
speech recognition to the year 2030. The predictions show that the coming
decade could be just as exciting and important to the development of speech
recognition and spoken language understanding as the previous one. We still
have many research problems to solve before speech recognition will reach the
point where it works all the time, for everyone. However, this is a goal worth
working toward, as speech recognition is a key component to more fluid,
natural, and accessible interactions with technology.

\section*{\large Acknowledgements}

Thanks to Chris Lengerich, Marya Hannun, Sam Cooper, and Yusuf Hannun for their
feedback on this article.

\bibliographystyle{plainnat}
\bibliography{references}

\end{document}